\documentclass[runningheads]{llncs}

\usepackage{eccv}

\usepackage{eccvabbrv}

\usepackage{graphicx}
\usepackage{booktabs}
\usepackage{microtype}

\usepackage[accsupp]{axessibility}  

\usepackage[pagebackref,breaklinks,colorlinks]{hyperref}

\usepackage{orcidlink}

\newcommand{\bA}{\mathbf{A}}

\newcommand{\bD}{\mathbf{D}}

\newcommand{\bff}{\mathbf{f}} 

\newcommand{\bI}{\mathbf{I}}

\newcommand{\bn}{\mathbf{n}}\newcommand{\bN}{\mathbf{N}}

\newcommand{\bp}{\mathbf{p}}

\newcommand{\bS}{\mathbf{S}}
\newcommand{\bt}{\mathbf{t}}
\newcommand{\bu}{\mathbf{u}}
\newcommand{\bv}{\mathbf{v}}\newcommand{\bV}{\mathbf{V}}

\newcommand{\bx}{\mathbf{x}}

\newcommand{\bpi}{\boldsymbol{\pi}}

\newcommand{\bPhi}{\boldsymbol{\Phi}}

\newcommand{\cG}{\mathcal{G}}

\newcommand{\cI}{\mathcal{I}}

\newcommand{\cL}{\mathcal{L}}

\newcommand{\cP}{\mathcal{P}}

\newcommand{\cX}{\mathcal{X}}

\newcommand{\figref}[1]{Figure~\ref{#1}}
\newcommand{\secref}[1]{Section~\ref{#1}}

\newcommand{\eqnref}[1]{Eq.~\ref{#1}}
\newcommand{\tabref}[1]{Table~\ref{#1}}

\newcommand{\abs}[1]{\left\lvert#1\right\rvert}

\newcommand{\boldstartspace}[1]{\vspace{0.1in}\noindent\textbf{#1}\hspace{0.5em}}

\newcommand{\RRR}{\mathbb{R}}

\begin{document}

\title{LaRa: Efficient Large-Baseline Radiance Fields}




\author{{Anpei Chen\textsuperscript{1,2}\orcidlink{0000-0003-2150-2176} \quad Haofei Xu\textsuperscript{1,2}\orcidlink{0000-0003-1313-3358} \quad Stefano Esposito\textsuperscript{1}\orcidlink{0000-0001-7624-1984}\\ Siyu Tang\textsuperscript{2}\orcidlink{0000-0002-1015-4770} \quad Andreas Geiger\textsuperscript{1}\orcidlink{0000-0002-8151-3726}}}

\institute{
{\textsuperscript{1}University of T\"ubingen, T\"ubingen AI Center}
\quad
{\textsuperscript{2}ETH Z\"urich}\vspace{8pt}\\
 \url{https://apchenstu.github.io/LaRa/}
}


\authorrunning{Chen et al.}



\maketitle

{
    \centering
    \captionsetup{type=figure}
    \includegraphics[width=\textwidth]{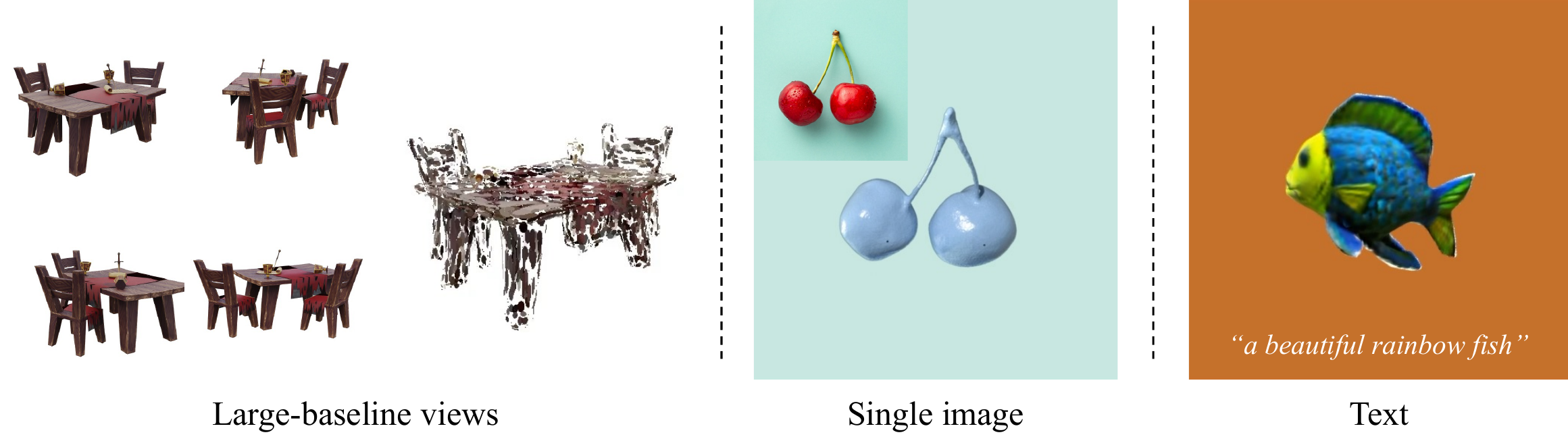}
    \captionof{figure}{LaRa is a feed-forward 2D Gaussian Splatting model that reconstructs radiance fields from large-baseline views, a single image, or a text prompt.}
    \label{fig:teaser}
}

\begin{abstract}
Radiance field methods have achieved photorealistic novel view synthesis and geometry reconstruction.
But they are mostly applied in per-scene optimization or small-baseline settings.
While several recent works investigate feed-forward reconstruction with large baselines by utilizing transformers, they all operate with a standard global attention mechanism and hence ignore the local nature of 3D reconstruction.
We propose a method that unifies local and global reasoning in transformer layers, resulting in improved quality and faster convergence. Our model represents scenes as Gaussian Volumes and combines this with an image encoder and Group Attention Layers for efficient feed-forward reconstruction. Experimental results demonstrate that our model, trained for two days on four GPUs, demonstrates high fidelity in reconstructing 360$^{\circ}$ radiance fields, and robustness to zero-shot and out-of-domain testing. 


\keywords{3D Reconstruction \and 3D Transformer \and Radiance Fields}
\end{abstract}

\section{Introduction}
\label{sec:intro}

The ability to reconstruct the shape and appearance of objects from multi-view images has long been one of the core challenges for computer vision and graphics.
Modern 3D reconstruction techniques achieve impressive results with various applications in visual effects, e-commerce, virtual and augmented reality, and robotics.
However, they are limited to small camera baselines or dense image captures~\cite{wang2021ibrnet,Chen2021ICCVb,Mildenhall2020ECCV,kerbl3Dgaussians}.
In recent years, the computer vision community has made great strides towards high-quality scene reconstruction.
In particular, Structure-from-Motion~\cite{Schoenberger2016CVPR,NOAH2006TOG} and multi-view stereo \cite{yao2018mvsnet,Goesele2007ICCV} emerged as powerful 3D reconstruction methods.
They identify surface points by aggregating similarities between point features queried from source images, and are able to reconstruct highly accurate surface and texture maps.

Despite these successes, geometry with view-consistent textures is not the only aspect required in applications of 3D reconstruction.
The reconstruction process should also be able to recover view-dependent appearance.
To this end, neural radiance fields \cite{Mildenhall2020ECCV} and neural implicit surfaces \cite{Niemeyer2020CVPR,Yariv2020NIPS} investigate volumetric representations that can be learned from multi-view captures without explicit feature matching.
Their follow-ups \cite{Wang2021NEURIPSa,Yu2021CVPR,Mueller2022TOG,Keil2022CVPR,Chen2022ECCV,kerbl3Dgaussians,SunSC22,barron2021mipnerf,Yu2023MipSplatting} improve efficiency and quality, but mainly require per-scene optimization and dense multi-view supervision.

Several recent works thus investigate feed-forward models for radiance field reconstruction while relaxing the dense input view requirement.
While feed-forward designs vary, they commonly utilize local feature matching \cite{Chen2021ICCVb,SRF,GeoNeRF,ENeRF,xu2023murf}, which however limits them to
small-baseline reconstruction, since feature matching generally relies on substantial image overlap and reasonably similar viewpoints.
Geometry-aware transformers \cite{reizenstein21co3d,kulhanek2022viewformer,venkat2023geometry,Miyato2024GTA} have also been adapted to address large-baseline problems, but they often suffer from blurry reconstructions due to the lack of 3D inductive biases.
Recent large reconstruction models \cite{hong2024lrm,li2024instantd} learn the internal perspective relationships through context attention, enabling large-baseline reconstruction.
However, the transformers are unaware of epipolar constraints, and instead are tasked to implicitly learn spatial relationships, which requires substantial data and GPU resources.

In this work, we present \emph{LaRa}, a feed-forward reconstruction model without the requirement of heavy training resources for the task of 360$^{\circ}$ bounded radiance fields reconstruction from unstructured few-views.
The core idea of our work is to progressively and implicitly perform feature matching through a novel volume transformer.
We propose a Gaussian volume as the 3D representation, in which each voxel comprises a set of learnable Gaussian primitives.
To obtain the Gaussian volume from image conditions, we progressively update a learnable embedding volume by querying features in 3D.
Specifically, we utilize a DINO image feature encoder to obtain image tokens and lift 2D tokens to 3D by unprojecting them into a shared canonical space. 
%
%
%
Next, we propose a novel \emph{Group Attention Layer} architecture to enable local and global feature aggregation. 
Specifically, we divide dense volumes into local groups and only apply attention within each group, inspired by standard feature point matching. 
The grouped features and embeddings are fed to a cross-attention sub-layer to implicitly match features between feature groups of the feature volume and embedding volume, 
which is followed by a 3D CNN layer to efficiently share information across neighboring groups.
After passing through all attention layers, the volume transformer outputs a Gaussian volume, and is then decoded as 2D Gaussian~\cite{HUANG2024SIGGRAPH} parameters using a coarse-to-fine decoding process.
By incorporating efficient rasterization, our method achieves high-resolution renderings.


%

We demonstrate our method's efficiency and robustness for providing photorealistic, 360$^\circ$ novel view synthesis results using only four input images.
%
%
We find that our model achieves zero-shot generalization to significantly out-of-distribution inputs.
Moreover, our reconstructed radiance fields allow high-quality mesh reconstruction using off-the-shelf depth-map fusion algorithms. Finally, our model achieves high-quality reconstruction results using only 4 A100-40G GPUs within a span of 2 days.

\section{Related Work}

\boldstartspace{Multi-view stereo.}
Multi-view stereo reconstruction aims to generate detailed 3D models by reasoning from images captured from multiple viewpoints, which has been studied for decades \cite{de1999poxels,kutulakos2000theory,kolmogorov2002multi,esteban2004silhouette,seitz2006comparison,furukawa2010accurate,schonberger2016pixelwise}.
In recent years, multi-view stereo networks \cite{yao2018mvsnet,im2018dpsnet} have been proposed to address MVS problems.
MVSNet \cite{yao2018mvsnet} utilizes a 3D Convolutional Neural Network for processing a cost volume.
This cost volume is created by aggregating features from a set of adjacent views, employing the plane-sweeping technique from a reference viewpoint, facilitating depth estimation and enabling superior 3D reconstructions.
Subsequent research has built on top of this foundation, incorporating strategies such as iterative plane sweeping \cite{yao2019recurrent}, point cloud enhancement \cite{chen2019point}, confidence-driven fusion \cite{luo2019p}, and the usage of multiple cost volumes \cite{cheng2020deep,gu2020cascade} to further refine reconstruction accuracy.
 However, all of these works require a large image overlap to achieve faithful feature matching.

\boldstartspace{Few-shot Radiance fields.}
The Radiance field representation \cite{Mildenhall2020ECCV} has revolutionized the reconstruction field, emerging as a promising replacement for traditional reconstruction methods.
Despite the promising achievement in per-scene sparse view reconstruction \cite{Niemeyer2022CVPR,SparseNeRF,kangle2021dsnerf,somraj2022decompnet,sparf2023,Chen2023TOG,Chen2023factor,shi2023zerorf,wu2023reconfusion}, training a feed-forward radiance field predictor \cite{Yu2021CVPR,wang2021ibrnet,Chen2021ICCVb,SRF} has gained popularity.
MVSNeRF \cite{Chen2021ICCVb} proposed to combine a cost volume with volume rendering, allowing appearance and geometry reconstruction only using a photometric loss.
The following works \cite{GeoNeRF,ENeRF,chen2023matchnerf,xu2023murf} are proposed to advance reconstruction quality and efficiency.
Similarly to standard MVS methods, they are limited to small camera baselines.

Recently, several works have explored feed-forward models for few-shot \cite{anciukevicius24iclr,li2024instantd,du2023cross,Miyato2024GTA,charatan23pixelsplat,chen2024mvsplat,dust3r2023} input by capitalizing on large-scale training datasets and model sizes.
They leverage cross-view attention to global reason about 3D scenes and output 3D representation (e.g., tri-plane, IB-planes) for radiance field reconstruction.
Concurrent work by LGM \cite{tang2024lgm} and GRM \cite{xu2024grm} introduces few-shot 3D reconstruction models that produce high-resolution 3D Gaussians using a transformer framework.
While these methods achieve impressive visual results, training becomes expensive and less practical for the academic community.
Unlike some recent single view reconstruction methods~\cite{szymanowicz2024splatter_image,hong2024lrm}, our work focuses on few-shot ($>1$) reconstruction since single-view input can be efficiently lifted to multi-view by multi-view generative models \cite{liu2023zero1to3,xu2024dmvd,shi2023MVDream,long2023wonder3d}.

\section{LaRa: Large-baseline Radiance Fields}

Our goal is to reconstruct the geometry and view-dependent appearance of bounded scenes from sparse input views using limited training resources.
Given $M$ images $\bI \!=\! (I_1, \ldots, I_M)$ with camera parameters $\bpi \!=\! (\pi_1, \ldots, \pi_M)$,
our method reconstructs radiance fields as a collection of 2D Gaussians, which is used to synthesize novel views and extract meshes.
Our model is a function $\bff$ of a discrete radiance field of voxel positions $\bv$ and produces a Gaussian volume $\bV_{\cG}$

\begin{align}
\bV_{\cG} = \{\cG^k_i\}_{k=1}^K = \bff(\bv; \bI,\bpi) \text{,}
\end{align}
%
where $\cG^k_i$ represents the primitives within $i$th voxel, and $k$ is the index of $K$ primitives.
The output Gaussian volume $\bV_{\cG}$ can be utilized for decoding into radiance fields.
Our work considers sparse input views, in which the camera rotates around a bounded region within a hemisphere.
Our approach is designed to handle unstructured views and is flexible to accommodate various numbers of views (see supplementary material).
\figref{fig:pipeline} shows an overview of our method.

In the following, we first describe how we model objects using Gaussian Volumes,
in which each voxel stores multiple Gaussian primitives (\secref{sec:representation}).
Next, we introduce how to infer the primitive parameters from multi-view inputs (\secref{sec:feedforward}).
For rendering, we explore a coarse-fine decoding process to enable efficient rendering with rich texture details (\secref{sec:renderer}).
Finally, we discuss how we train our model from large-scale image collections (\secref{sec:training}).

\begin{figure}[t]
    \centering
    \includegraphics[width=0.98\linewidth]{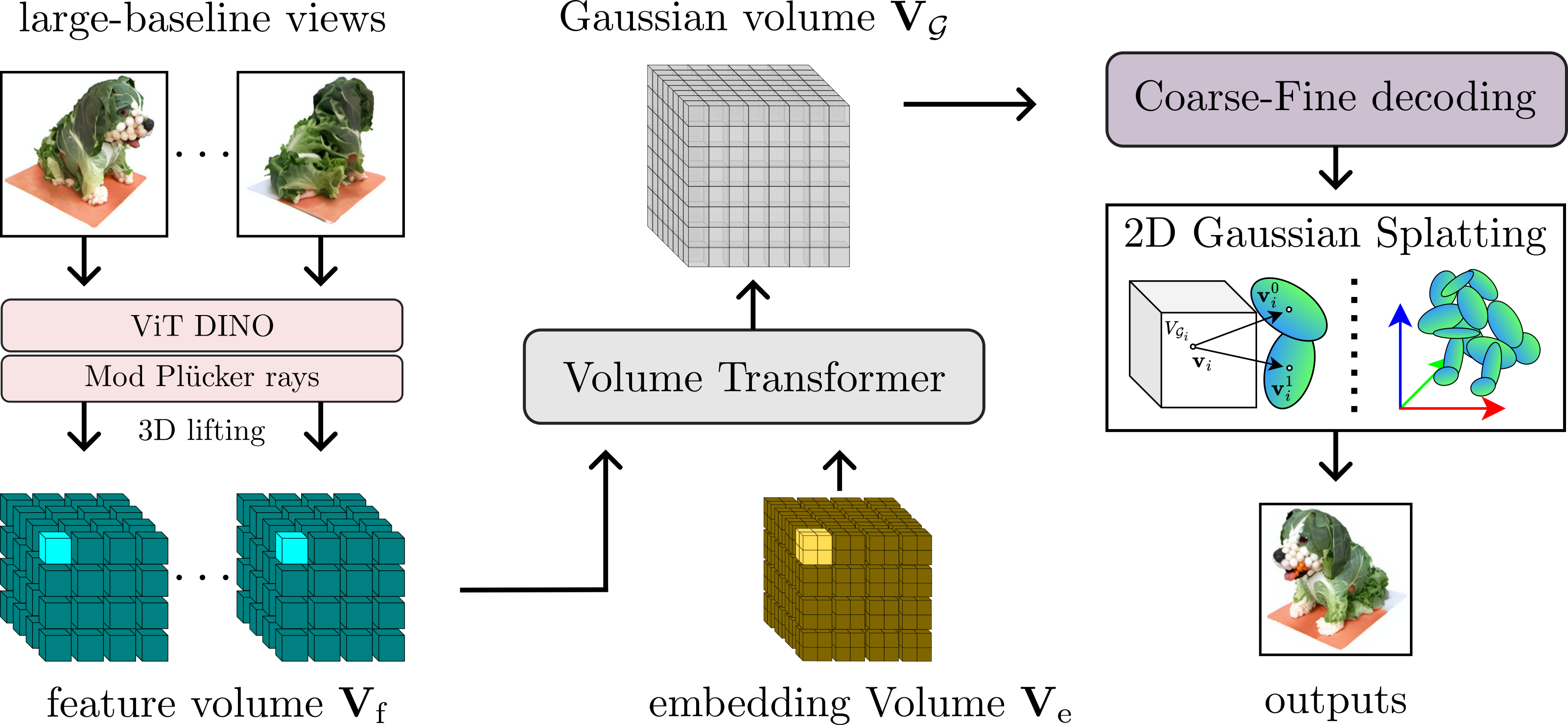}
    \caption{\textbf{Pipeline.}
        Our method represents objects as dense voxels filled with 2D Gaussian primitives.
        We first construct 3D feature volumes $\bV_\text{f}$ by lifting 2D DINO features to a canonical volume, modulated by Plücker rays (\secref{sec:representation}).
        We then apply a volume transformer to reconstruct a Gaussian volume $\bV_\cG$ from the feature and embedding volumes (\secref{sec:feedforward}).
        We use a coarse-to-fine decoding process to regress 2D Gaussian primitive parameters (\secref{sec:renderer}), followed by rasterization for efficient rendering.}
    \label{fig:pipeline}
\end{figure}

\subsection{3D Representation}
\label{sec:representation}
We utilize a 3D voxel grid as our 3D representation, consisting of 3 volumes: an \textit{image feature volume} $\bV_\text{f}$ to model image conditions, an \textit{embedding volume} $\bV_\text{e}$ describes 3D prior learned from data, and a \textit{Gaussian volume} $\bV_{\!\cG}$ represents the radiance field.

\boldstartspace{Image feature volume.}
We construct a feature volume for each input view by lifting the 2D image features to a canonical volume defined in the center of the scene.
We use the DINO \cite{Caron2021ICCV} image encoder to extract per-view image features, and inject Plücker ray directions into the features via the adaptive layer norm \cite{DiT}.
Unlike previous works that modulate camera poses to image features using extrinsic and intrinsic matrices~\cite{liu2023zero1to3,hong2024lrm}, Plücker rays are defined by the cross product between the camera location and ray direction, offering a unique ray parameterization independent of object scale, camera position, and focal length.
%
After modulation, we obtain $M$ per-view image feature maps. 
We further lift the 2D maps to 3D by back-projecting the feature maps to a canonical volume, therefore resulting in $M$ feature volumes $\bV_\text{f} \in \RRR^{W \times W \times W \times O}$ with $O$ channels.


\boldstartspace{Embedding volume.}
Inspired by prior works \cite{Karras2019CVPR,Nguyen-Phuoc2019ICCV,hong2024lrm}, we construct a learnable embedding volume $\bV_\text{e} \in \RRR^{W \times W \times W \times C}$ to model prior knowledge.
3D reconstruction is generally under-constrained in sparse view settings, hence prior knowledge is critical for faithful reconstructions.
We propose to leverage a 3D embedding volume to model and learn prior information across objects, which acts as a 3D object template that greatly reduces the solution space.
The embedding volume is aligned with the image feature volume, allowing for efficient cross attention (see \secref{sec:feedforward}).

\boldstartspace{Gaussian volume.}
To achieve efficient rendering, we propose to use dense primitives as an object representation and output a set of 2D Gaussians from the image feature volume and embedding volume.
However, predicting a set of dense unordered point sets without 3D supervision is always a challenge for neural networks.
To this end, we introduce a dense Gaussian volume representation that can effectively model points densely near the object's surface, while being suitable for modern network architectures by facilitating prediction and generation.

Specifically, our Gaussian volume comprises $K$ learnable Gaussian primitives per voxel, where each primitive can move freely within a constrained spherical region centered at the voxels' center.
For primitive modeling, we borrow the shape and appearance parametrization from 2D Gaussian splatting \cite{HUANG2024SIGGRAPH} for better surface modeling. 
Each Gaussian has an opacity $\alpha$, tangent vectors $\bt \!=\! [\bt_\text{u},\bt_\text{v}]$, a scaling vector $\bS \!=\! (s_\text{u},s_\text{v})$ controlling the shape of the 2D Gaussian, and spherical harmonics coefficients for view-dependent appearance.
Furthermore, we substitute the primitive's position with an offset vector $\Delta \!\in\! [-1,1]^3$, incorporating a scaled sigmoid activation function.
Consequently, the position of Gaussian primitive $k$ in voxel $\bv_i$ is expressed as $\bp^k_i = \bv_i + r \cdot \Delta^k_i$, where $r$ signifies the maximum displacement range of the primitive.
In this way, primitives are restricted to neighborhoods of uniformly distributed local centers.
The inclusion of offset modeling allows each voxel to effectively represent adjacent regions that require it.
This reduces unnecessary capacity in empty space and enhances the representational capacity compared to the standard dense volume.
We refer the reader to the supplementary material for more details on 2D Gaussian splatting.

\subsection{Volume Transformer}
\label{sec:feedforward}

To predict the Gaussian volume, we propose a volume transformer architecture to perform attention between volumes.
Self-attention and cross-attention modules, as commonly used in transformers \cite{Transformers}, are inefficient for volumes, since the number of tokens grows cubically with the resolution of the 3D representation.
Na\"ive applications thus result in long training times and large GPU memory requirements.
In addition, geometry constraints and regional matching play crucial roles in the context of 3D reconstruction, which should be considered in the attention design.


\begin{figure}[t]
    \centering
    \includegraphics[width=0.98\linewidth]{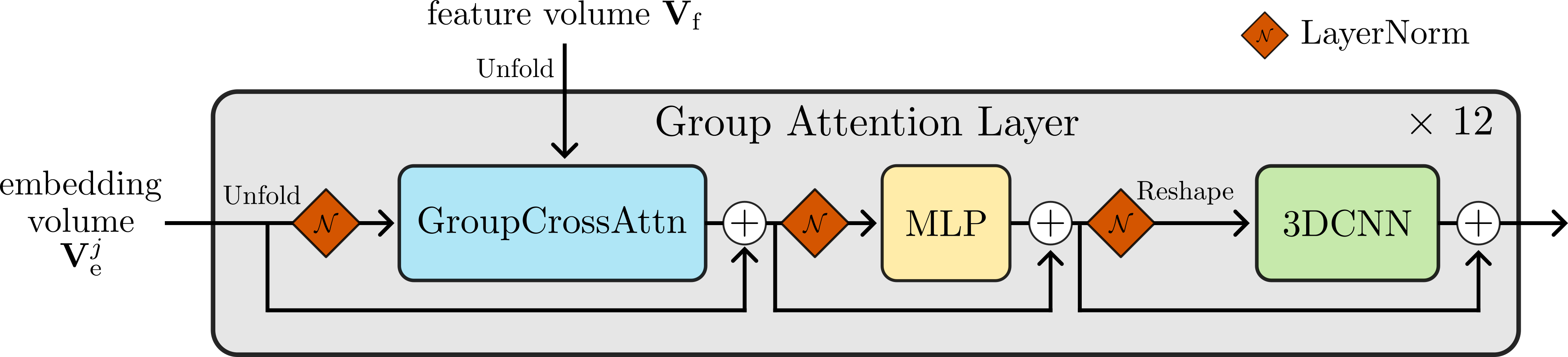}
    \caption{\label{fig:vol_transformer}%
    \textbf{Volume Transformer.}
    We aggregate the embedding volume $\bV_\text{e}$ and feature volume $\bV_\text{f}$ through a series of Group Attention Layers that progressively match features.
    In each layer, the volumes are first unfolded into local groups.
    Subsequently, a layer normalization is applied, followed by a GroupCrossAttn sublayer.
    This is followed by another normalization and an MLP layer.
    The output is reshaped back to the original embedding volume shape, processed by a 3D convolution layer, and forwarded to the next layer.
    To connect the output of the sublayers, we use
    residual connections.
}
\end{figure}

We now present our novel volume transformer containing a set of \emph{group attention layers} that progressively update the embedding volume.
%
Our group attention layers contain three sublayers (see \figref{fig:vol_transformer}): group cross-attention, a multi-layer perceptron (MLP), and 3D convolution.
Given the image feature volume and embedding volume, we first unfold these 3D volumes (i.e., $\bV_\text{f}$ and $\bV_\text{e}$) into $G$ groups along each axis.
We then apply a cross-attention layer between the corresponding groups of embedding tokens $\bV_\text{e}^{g,j}$ and image feature tokens $\bV^{g}_\text{f}$, where $j$ denotes the index of the layer starting from 1, and $\{\bV_\text{e}^{g,1}\}_g \!=\! \bV_\text{e}$.
\figref{fig:pipeline} illustrates the unfolding of $G=4$ and highlights the corresponding groups.

The next sublayer is
an MLP, similar to the original transformer \cite{Vaswani2017NIPS,DiT,Perceiver}.
The updated embedding groups $\{ \ddot{\bV}_\text{e}^{g,j} \}^G_{g=1}$ are reassembled into the original volume shape, resulting in $\ddot{\bV}_\text{e}^{j}$, which are subsequently processed by a 3D convolutional layer to 
share information between groups and enable the intra-model connections within the spatially organized voxels.
In summary,

\begin{align}
\dot{\bV}_\text{e}^{g,j} &= \text{GroupCrossAttn}\left(\text{LN}\left(\bV_\text{e}^{g,j}\right), \bV^{g}_\text{f}\right) + \bV_\text{e}^{g,j} \text{,} \\
\ddot{\bV}_\text{e}^{g,j} &= \text{MLP}\left(\text{LN}\left(\dot{\bV}_\text{e}^{g,j}\right)\right) + \dot{\bV}_\text{e}^{g,j} \text{,} \\
\bV_\text{e}^{j+1} &= \text{3DCNN}\left(\text{LN}\left(\ddot{\bV}_\text{e}^{j}\right)\right) + \ddot{\bV}_\text{e}^{j} \text{.}
\end{align}

To incorporate information from multiple views, we flatten and concatenate the image feature tokens from multi-view feature volumes.
It is important to note that different groups are processed simultaneously by the group attention layer across the batch dimension.
This parallel processing allows for a larger training batch size within the attention sublayer, reducing the number of training steps required.
%
In addition, using a 3D convolution layer increases inference efficiency compared to the popular self-attention layer. Also, we also apply layer norms $\text{LM}(\cdot)$ between the sub-layers. 
Finally, the output embedding volume $\bV^j_\text{e}$ serves as input for the subsequent $(j\!+\!1)$\textsuperscript{th} group attention layer. 

After passing through all (12 in our experiments) group attention layers, we use a 3D transposed CNN to scale up the updated embedding volume $\dot{\bV}_\text{e}$,
%
$
\bV_{\!\cG} = \text{Transpose-3DCNN}\left(\dot{\bV}_\text{e}\right) \!\text{.}
$
%
Now we have a Gaussian volume $\bV_{\cG}$, each Gaussian voxel is a 1-D feature vector $\bV^i_{\cG} \in \RRR^{1\times B}$, representing the primitives associated with the voxel.

\subsection{Coarse-Fine Decoding}
\label{sec:renderer}

We obtain 2D Gaussian primitive shape and appearance parameters from the Gaussian volume, so we introduce a coarse-fine decoding process to better recover texture details. 
Instead of using a single network and sampling scheme to reason about the scene, we simultaneously optimize two decoding modules: one ``coarse'' and one ``fine''. Intuitively, the ``fine'' decoding module attempts to learn a geometry-aware texture blending process based on multi-view images, primitive features, and rendering buffers from the coarse module.

\begin{figure}[t]
    \centering
    \includegraphics[width=\linewidth]{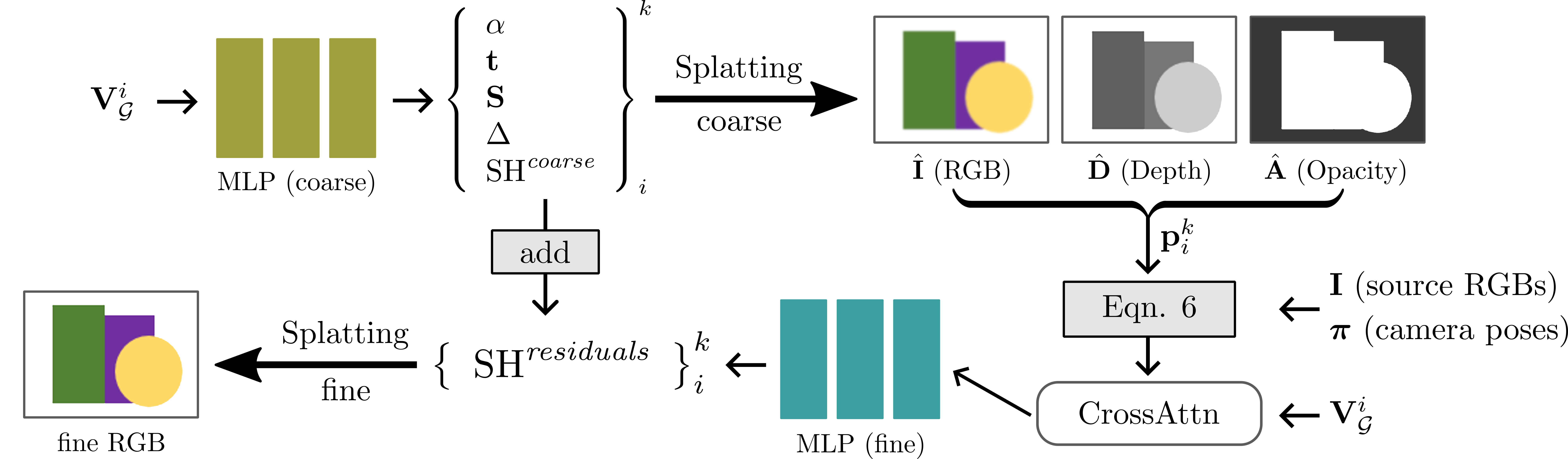}
    \caption{\textbf{Coarse-fine decoding.} Top row: A ``coarse'' decoding module transforms the voxel features $\mathbf{V}^i_{\mathcal{G}}$ into $K$ 2D Gaussian parameters, representing shape (specifically, $\alpha, \mathbf{t}, \mathbf{S}, \Delta$) and appearance (denoted as $\text{SH}^{\textit{coarse}}$). This step is followed by a splatting procedure.
    On the bottom, a "fine" decoding module aggregates rendering buffers (i.e., RGB, depth, and alpha maps) from the coarse module, volume feature, and source images for appearance enhancement.
    It projects the centers of primitives onto these buffers, applies cross-attention with the voxel features $\mathbf{V}^i_{\mathcal{G}}$, and produces residual spherical harmonics $\text{SH}^{\textit{residuals}}$. These residuals are added to the coarse spherical harmonics for a refined splatting process.
    }
    \label{fig:decoding}
\end{figure}

For the ``coarse'' decoding module, we feed Gaussian volume features to a lightweight MLP and output a set of $K$ Gaussian parameters per voxel.
We employ the efficient 2D splatting technique \cite{HUANG2024SIGGRAPH} to form high-resolution renderings, including RGB, depth, opacity, and normal maps.
During training, we render $M$ input views and $M$ novel views for supervision.

Despite the fact that the coarse renderings can already provide faithful depths/geometries, the appearance tends to be blurred, as shown in (e) of ~\figref{fig:ablation}.
This is because the image texture can easily be lost after the DINO encoder and the Group Attention layers.
To address this problem, we propose a ``fine'' decoding module to guide the fine texture prediction.

Specifically, we project the primitive centers $\bp^k_i$ onto the coarse renderings (i.e., RGB image $\hat{\bI}$, depth image $\hat{\bD}$, and accumulation alpha map $\hat{\bA}$) to contain the coarse renderings for each primitive using the camera poses $\bpi$,
\begin{align}
\cX_{\bp^k_i} = \left(\bI_{\bp^k_i},\hat{\bI}_{\bp^k_i}, \hat{\bD}_{\bp^k_i}, \hat{\mathrm{\bA}}_{\bp^k_i}\right) = \bPhi \left(\cP \left(\bp^k_i, \bpi \right), \oplus \left[\bI,\hat{\bI}, \hat{\bD}, \hat{\mathrm{\bA}}\right] \right) \text{,}
\end{align}
where $\cP$ denotes the point projection, $\oplus$ is a concatenation operation along the channel dimension, and $\bPhi$ is a bilinear interpolation in 2D space.

In practice, the depth features can change significantly in different scenes.
To mitigate scaling discrepancies, we replace the rendering depth $\hat{\bD}_{\bp^k_i}$ with a displacement feature $\abs{\hat{\bD}_{\bp^k_i} - z_{\bp^k_i}}$ that compares the rendered depth for input views and the depth $z_{\bp^k_i}$ of a primitive, allowing for occlusion-aware reasoning.

We then apply a point-based cross-attention layer to establish relationships between the features of a point $\cX_{\bp^k_i}$ and the primitive voxel.
The results of this cross-attention process are then fed into an MLP, which is tasked with predicting the residual  spherical harmonics
\begin{align}
\text{SH}_{i,k}^\textit{residuals} &= \text{MLP}\left(\text{CrossAttn}\left(\cX_{\bp^k_i}, \bV^i_{\mathcal{G}} \right) \right) \text{,} \\
\text{SH}_{i,k}^\textit{fine} &= \text{SH}_{i,k}^\textit{coarse} + \text{SH}_{i,k}^\textit{residuals} \text{.}
\end{align}
Furthermore, both coarse and fine modules are differentiable and updated simultaneously.
Thus, the fine renderings can further regularize the coarse predictions.

\boldstartspace{Splatting.}
\label{sec:splatting}
Our work takes advantage of Gaussian splatting \cite{kerbl3Dgaussians,HUANG2024SIGGRAPH} to facilitate efficient high-resolution image rendering.
We follow the original rasterization process and further output depth and normal maps by integrating the $z$ value and the normal of the primitives.

\subsection{Training}
\label{sec:training}

Our LaRa is optimized across scenes via gradient descent, minimizing simple image reconstruction objectives between the coarse and fine renderings (i.e., $\hat{\cI}$) and the ground-truth images (i.e., $\cI$),
\begin{align}
\cL = \cL_\text{MSE}(\cI,\hat{\cI}) + \cL_\text{SSIM}(\cI,\hat{\cI}) + \cL_\text{Reg} \text{,}
\label{eq:loss}
\end{align}
where $\cL_\text{MSE}$ is the pixel-wise L2 loss, $\cL_\text{SSIM}$ is the structural similarity loss, which are applied on both coarse and fine RGB outputs.

\boldstartspace{Regularization terms.}
We find that only applying the photometric reconstruction losses is adequate for rendering.
However, the consistency across views is low because of the strong flexibility of the discrete Gaussian primitives.
To encourage the primitives to be constructed on the surface, we follow 2D Gaussian splatting~\cite{HUANG2024SIGGRAPH} that utilize a self-supervised distortion loss $\cL_\text{d}$ and a normal consistency loss $\cL_\text{n}$ to regularize the training.

Specifically, we concentrate the weight distribution along the rays by minimizing the distance between the ray-primitive intersections, inspired by Mip-NeRF\cite{barron2021mipnerf}.
Given a ray $\bu(\bx)$ of pixel $\bx$, we obtain its distortion loss by, 
\begin{equation}
\cL_\text{d} = \sum_{i,j}\omega_i\omega_j|z_i-z_j|\text{,}
\label{eqn:loss_distortion}
\end{equation}
where $\omega_i = \alpha_i\,\cG_i(\bu(\bx))\prod_{j=1}^{i-1} (1 - \alpha_j\,\cG_j(\bu(\bx)))$ is the blending weight of the $i-$th intersection and $z_i$ the depth of the intersection point.

As 2D Gaussians explicitly model the primitive normals, we can align their normals $\bn_i$ with the normals $\bN$ derived from the depth maps via the loss
\begin{align}
\cL_\text{n} = \sum_{i} \omega_i (1-\bn_i^\top\bN)\text{.}
\label{eqn:loss_normal}
\end{align}
Therefore, our regularization term for the ray $\bu(\bx)$ is given by $\cL_\text{Reg} = \gamma_\text{d} \cL_\text{d} + \gamma_\text{n} \cL_\text{n}$.
We set $\gamma_\text{d} \!=\! 1000$ and $\gamma_\text{n} \!=\! 0.2$ in our experiments.

\section{Implementation Details}
We briefly discuss our implementation, including the training and evaluation dataset, network design, optimizer, and mesh extraction.

\boldstartspace{Datasets.}
We train our model on multi-view synthetic renderings of objects \cite{gobjaverse, qiu2024richdreamer}, based on the Objaverse dataset \cite{Objaverse}, which includes 264,775 scenes with a train/test split of 10:1.
Each scene contains 38 circular views with an image resolution of $512 \times 512$.
To ensure sufficient angular coverage of the input views, we employ the classical K-means algorithm to cluster the cameras into 4 clusters.
We employ eight views for supervision for each object and leverage the loss objectives outlined in \eqnref{eq:loss} to update the network.

We present our in-domain evaluation using the Objaverse dataset's test set, consisting of 26,478 scenes.
To assess our model's cross-domain applicability, we conducted tests on the Google Scanned Objects dataset \cite{GSO}, which contains 1,030 scans of real objects,
and on the 46 hydrants and 90 teddy bears from the Co3D test set \cite{reizenstein21co3d}, totaling 136 objects.
To examine our model's performance on zero-shot reconstruction task, we use the generative multi-view dataset from Instant3D~\cite{li2024instantd}, which comprises 122 scenes generated from text prompts.

\boldstartspace{Network.}
We developed LaRa using PyTorch Lightning \cite{PyTorchLightning} and conducted our training on 4 NVIDIA A100-40G GPUs over a period of 2 days for the fast model and 3.5 days for the base model, with a batch size of 2 per GPU.
We use DINO-base for encoding $M \!=\! 4$ multi-view images at a resolution of $512 \times 512$.
We use a volume resolution of $W \!=\! 16$ with $C \!=\! 768$ channels for the image feature volume,
and a resolution of $W \!=\! 32$ with $C \!=\! 256$ channels for the embedding volume, dividing both into $G \!=\! 16$ groups for the group attention layers.
Our group attention network consists of 12 layers, producing a Gaussian volume of size $64 \!\times\! 64 \!\times\! 64 \!\times\! 80$.
We choose $K \!=\! 2$ primitives for each voxel, and constrain the offset radius to $r=1/32$ of the length of the bounding box. The total number of trainable parameters is 125 million.

\boldstartspace{Training.}
The optimization is carried out using the AdamW optimizer \cite{Loshchilov2019ICLR}, starting with a learning rate of $2 \times 10^{-4}$ and following a cosine annealing schedule with a period of 10 epochs.
Our final model is trained for 50 epochs, comprising 50,000 iterations for each epoch.
We observe that applying the regularization loss from the start can slow down the convergence regarding the shape.
This is because regularization objectives tend to encourage thinner surfaces, which may result in premature convergence to local minima if the shapes are noisy.
In our experiments, we thus enable regularization after the first 15 epochs.

\boldstartspace{Mesh extraction.}
To obtain a mesh from reconstructed 2D primitives, we generate RGBD maps by rendering along three circular video trajectories at elevations of 30°, 0°, and $-30$°.
Inside the scene bounding box, we construct a signed distance function volume and apply truncated SDF (TSDF) fusion to integrate the reconstructed RGB and depth maps, allowing for efficient textured mesh extraction.
In our experiments, we use a resolution of $256^3$ and set a truncation threshold of 0.02 for TSDF fusion.

\section{Experiments}
We now present an extensive evaluation of LaRa, our large-baseline radiance field.
We first compare with previous and concurrent works on in-domain and zero-shot generalization settings.
We then analyze the effect of local attention, regularization term, and renderer.

\subsection{Comparison}
\label{sec:comparison}
We compare our method against MVSNeRF \cite{Chen2021ICCVb}, MuRF \cite{xu2023murf}, and the concurrent work LGM \cite{tang2024lgm}.
The first two methods are key representatives of feature matching-based methods, and the latter shares a conceptually similar approach of using Gaussian primitives for large-baseline settings.
It is worth noting that while existing feed-forward radiance field reconstruction methods are capable of being evaluated in large-baseline settings, retraining these methods to establish a new large-baseline benchmark on the Objaverse dataset is both time and GPU intensive.
Here, we retrain MVSNeRF and the current state-of-the-art feed-forward radiance field reconstruction method MuRF \cite{xu2023murf}.

\begin{figure}
    \centering
    \includegraphics[width=\linewidth]{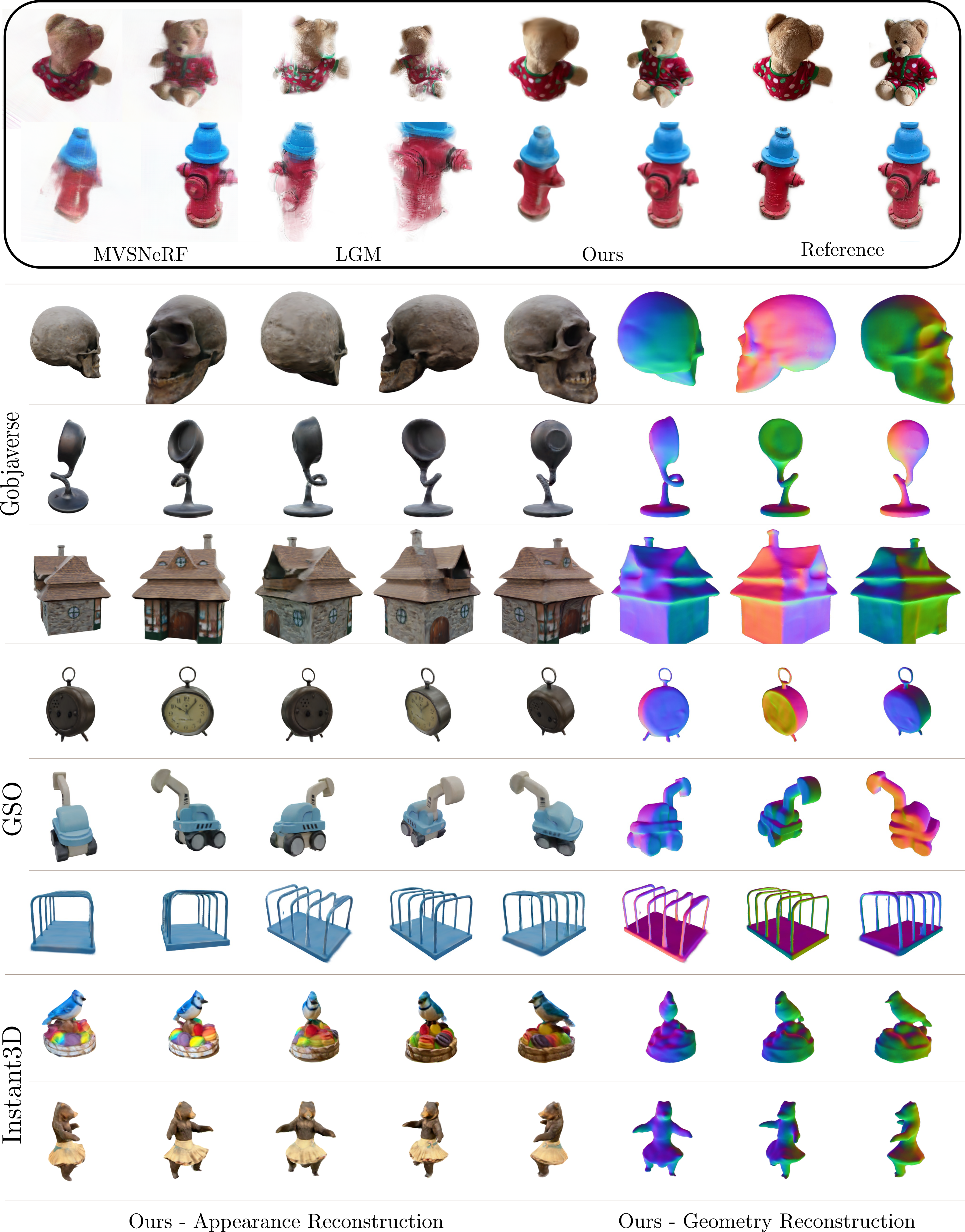}
    \caption{\textbf{Rendering results of unseen scenes.}
    The top two rows compare our reconstructions with MVSNeRF \cite{Chen2021ICCVb}, LGM \cite{tang2024lgm} on Co3D \cite{reizenstein21co3d}.
    We also show the view synthesis results for Gobjaverse \cite{gobjaverse}, GSO \cite{GSO}, and generative multi-view \cite{li2024instantd} datasets, arranged from top to bottom.
    Note that visual results from MuRF are not shown due to their lack of content, appearing as white images. The above results are reconstructed using 4 input views.
    }
    \label{fig:results}
\end{figure}

\begin{table}[t]
\caption{\textbf{Quantitative results of novel view synthesis.} Our fast model is trained for 30 epochs (2 days on 4 GPUs), while ours is trained for 50 epochs, taking 3.5 days.}
\vspace{-1mm}
\label{tab:app_comparison}
\centering
\setlength{\tabcolsep}{1.2pt}
\begin{tabular}{lccccccccc}
\toprule
& \multicolumn{3}{c}{Gobjaverse~\cite{gobjaverse}} & \multicolumn{3}{c}{GSO~\cite{GSO}} & \multicolumn{3}{c}{Co3D~\cite{reizenstein21co3d}} \\
\cmidrule(lr){2-4} \cmidrule(lr){5-7} \cmidrule(lr){8-10}
Method &  PSNR$\uparrow$ & SSIM$\uparrow$ & LPIPS$\downarrow$ & PSNR$\uparrow$ & SSIM$\uparrow$ & LPIPS$\downarrow$ &  PSNR$\uparrow$ & SSIM$\uparrow$ & LPIPS$\downarrow$\\
\midrule
MVSNeRF~\cite{Chen2021ICCVb}  & 14.48 & 0.896 & 0.1856 & 15.21 & 0.912 & 0.1544 & 12.94 & 0.841 & 0.2412\\%
MuRF~\cite{xu2023murf}  & 14.05 & 0.877 & 0.3018 & 12.89 & 0.885 & 0.2797 & 11.60 & 0.815 & 0.3933\\%
LGM~\cite{tang2024lgm}  & 19.67 & 0.867 & 0.1576 & 23.67 & 0.917 & 0.0637 & 13.81 & 0.739 & 0.4142\\%
Ours-fast      & 25.30 & 0.925 & 0.1027 & 26.79 & 0.946 & 0.0683 & 21.56 & 0.870 & 0.2079\\
Ours      & \textbf{26.14} & \textbf{0.931} & \textbf{0.0932} & \textbf{27.65} & \textbf{0.951} & \textbf{0.0616}  & \textbf{21.64} & \textbf{0.871} & \textbf{0.2026}\\
\bottomrule
\end{tabular}
\end{table}

\boldstartspace{Appearance.}
\tabref{tab:app_comparison} shows quantitative results (PSNR, SSIM, and LPIPS) comparisons.
Our method achieves clearly improved rendering quality for both in-domain generation (Gobjaverse testing set) and zero-shot generalization (GSO and Co3D datasets).
As shown in \figref{fig:results}, MVSNeRF fails to provide faithful reconstructions on the large-baseline setting and tends to produce floaters within the reconstruction regions since the cost volume is extremely noisy in the sparse view scenarios, resulting in a challenge for its convolution matching network to distinguish the surface.
MuRF \cite{xu2023murf} quickly overfits the white background and produces empty predictions for all inputs. 
Instead of predefining and constructing the feature similarity as network input, our method injects volume features to the inter-middle attention layer and implicitly and progressively matches them through the attention mechanism between the volume feature and updated embeddings, achieving clearer and overall better reconstructions.

As shown in \tabref{tab:app_comparison}, our approach is robust to scene scale and can be generalized to real captured images, such as those in the Co3D dataset, thanks to our canonical modeling and projection-based feature lifting.
In contrast, LGM \cite{tang2024lgm} leverages a monocular prediction and fusion technique that requires a reference scene scale and a constant camera-object distance to avoid focal length and distance ambiguity.
This requirement significantly limits its generalizability to real data.
As shown in \tabref{tab:app_comparison} and ~\figref{fig:results}, LGM provides faithful reconstructions in datasets with a strict constant camera-object distance, such as GSO, but fails to generalize to unconstraint multi-view data such as in Objaverse and Co3D datasets, and exhibits serious distortions.
Our model trained on 4 A100-40G GPUs for 2 days demonstrates superior results compared to the LGM model trained on 32 A100-80G GPUs (8$\times$ GPUs, 16$\times$ RAM, 32$\times$ GPU hours) and on the same synthetic Objaverse dataset \cite{Objaverse}.


Furthermore, our approach also performs well for generative multi-view images where textures are not consistent across views.
In this comparison, we only present a qualitative analysis due to the absence of ground truths, as illustrated in the bottom rows of~\figref{fig:results}. Our method offers detailed texture and smooth surface reconstruction. We invite the reader to our Appendix for more results.

\begin{figure}[t]
    \centering
    \includegraphics[width=\linewidth]{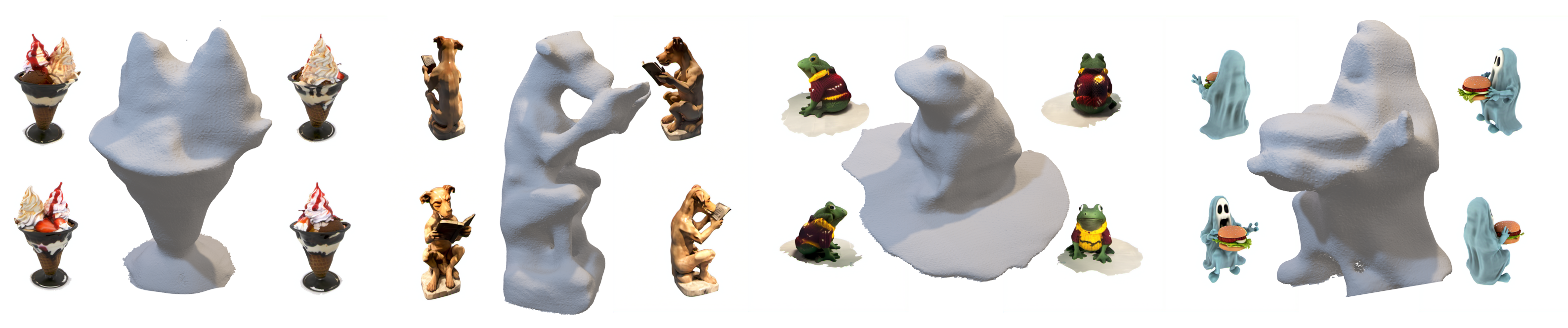}
    \caption{\textbf{Zero-shot reconstruction results.}
    Our approach achieves faithful surface reconstruction for images generated from text. These images are produced using a pre-trained text-to-image model~\cite{li2024instantd}. 
    }
    \label{fig:mesh}
\end{figure}

\begin{table}[t]
\caption{\textbf{Depth reconstruction.} We evaluate geometry quality within the mask by measuring  L1 error, and the accuracy in terms of the percentage of pixels that fall below thresholds of $[0.005,0.01,0.02]$.}
\vspace{-2mm}
\centering
\setlength{\tabcolsep}{6pt}
\begin{tabular}{lcccc}
\toprule
Method &  Abs err$\downarrow$ & Acc (0.005)$\uparrow$ & Acc (0.01)$\uparrow$ & Acc (0.02)$\uparrow$\\
\midrule
MVSNeRF~\cite{Chen2021ICCVb}  & 0.0993 & 6.2 & 12.4 & 24.0\\%
LGM~\cite{tang2024lgm}  & 0.1121 & 13.4 & 26.2 & 49.6\\
Ours-fast & 0.0695 & 32.7 & 52.2 & 70.7 \\
Ours & \textbf{0.0654} & \textbf{36.6} & \textbf{57.4} & \textbf{75.4} \\
\bottomrule
\end{tabular}
\label{tab:geo_comparison}
\end{table}

\boldstartspace{Geometry.}
We evaluate the quality of our geometry reconstruction by comparing the depth reconstructions on novel views, generated by a weighted sum of the $z$ values of the primitives.
As shown in \tabref{tab:geo_comparison}, our approach achieves significantly lower L1 errors and higher geometry accuracy than other baselines.
In \figref{fig:mesh}, we also visualize geometry reconstruction by extracting meshes using TSDF.
In addition, our trajectory video rendering (48 views at a resolution of $512 \time 512$) together with mesh extraction is highly efficient, as it does not require fine-tuning and can be performed in just 2 seconds.

\subsection{Ablation Study}
\label{sec:ablations}
We now analyze the contributions of individual elements of our model design.
%

\begin{figure}
    \centering
    \includegraphics[width=0.98\linewidth]{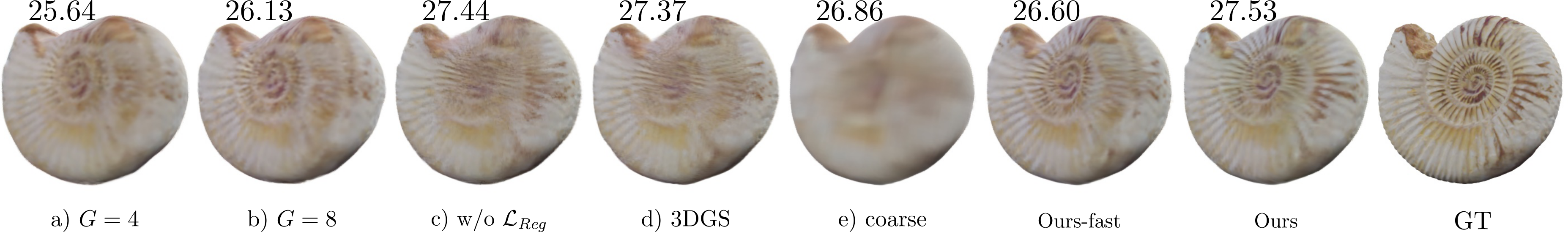}
    \caption{\textbf{Ablation study on a Shell scene.}
    We report the PSNR for each example at the top.
    Here, the fast model corresponds to the full model detailed in \tabref{tab:ablation}.}
    \label{fig:ablation}
\end{figure}

\boldstartspace{Effect of local attention.} 
We first evaluate the contribution of our group partition using different group numbers.
Here, $G \!=\! 1$ is equivalent to the standard cross-attention layer; however, using such group size can lead to much higher compute time for the same number of iterations, i.e., 22 days on 4 A100s for 30 epochs.
Therefore, our ablation starts with 4 groups for acceptable training time.
As shown in ablations (a), (b) and (g) in~\tabref{tab:ablation} and ~\figref{fig:ablation}, the image synthesis and geometry quality are consistently improved with a larger group number, thanks to the local attention mechanism. 

%

\begin{table}[t]
\caption{\textbf{Ablations.} See \secref{sec:ablations} for descriptions.}
\vspace{-2mm}
\centering
\setlength{\tabcolsep}{3.5pt}
\begin{tabular}{lccccccc}
\toprule
& \multicolumn{3}{c}{Gobjaverse~\cite{gobjaverse}} & \multicolumn{4}{c}{GSO~\cite{GSO}} \\
\cmidrule(lr){2-4} \cmidrule(lr){5-8}
Design &  PSNR$\uparrow$ & SSIM$\uparrow$ & LPIPS$\downarrow$ & PSNR$\uparrow$ & SSIM$\uparrow$ & LPIPS$\downarrow$ & Geo (\%) $\uparrow$ \\
\midrule

a) $G=4$  & 22.27 & 0.900 & 0.1558 & 23.06 & 0.920 & 0.1113 &  17.3/31.0/48.1\\
b) $G=8$  & 23.80 & 0.914 & 0.1256 & 25.30 & 0.936 & 0.0849 & 25.1/42.8/61.1\\
c) w/o $\cL_\textit{Reg}$ & \textbf{26.16} & \textbf{0.930} & \textbf{0.1006} & \textbf{27.71} & \textbf{0.950} & 0.0668 & 22.8/45.6/\textbf{71.2}\\
d) 3DGS & 26.04 & 0.929 & 0.1021 & 27.45 & 0.950 & \textbf{0.0666} & 23.3/45.0/69.2\\
e) coarse & 25.06 & 0.922 & 0.1239 &26.28 & 0.934 & 0.1017 & \textbf{32.7}/\textbf{52.2}/70.7 \\
f) SH order-0& 24.93 & 0.923 & 0.1097 & 26.71 & 0.945 & 0.0743 & 32.0/51.7/70.5\\
g) full model& 25.30 & 0.925 & 0.1027 & 26.79 & 0.946 & 0.0683 & \textbf{32.7}/\textbf{52.2}/70.7\\



\bottomrule
\end{tabular}
\label{tab:ablation}
\end{table}

\boldstartspace{Effect of regularization term.} 
We further evaluate the regularization term introduced in \eqnref{eqn:loss_distortion} and \eqnref{eqn:loss_normal}. We observe a marked improvement in the average rendering score when disabling the regularization. Although this provides a stronger model capability for modeling details, this may cause floaters near the surfaces, as shown in (c) and (d) of ~\figref{fig:ablation}, which leads to inconsistent free-viewpoint video rendering (see Appendix video). In contrast, our approach is able to reconstruct hard surfaces. 

\boldstartspace{Effect of renderer.}
We also compare 2D Gaussian splatting with 3D Gaussian splatting in our framework, as shown in  (c) and (d). They achieve similar rendering quality and we choose 2DGS to facilitate surface regularization and mesh extraction. Furthermore, to evaluate the effectiveness of the coarse-fine decoding, we conduct an evaluation of the coarse outputs, shown in row (e). Our fine decoding is able to provide richer texture details. 




\section{Conclusion}
We have presented \emph{LaRa}, a novel method for reconstructing 360$^{\circ}$ bounded radiance fields given large-baseline inputs. Our central idea is to match image features and embedding volume through unified local and global attention layers. By integrating this with a coarse-fine decoding and splatting process, we achieve high efficiency for both training and inference. In future work, we plan to explore how to enlarge the batch size per-GPU and volume resolution without increasing GPU usage. 







\bibliographystyle{splncs04}
\bibliography{bibliography_short,bibliography,bibliography_custom}

\begin{thebibliography}{10}
\providecommand{\url}[1]{\texttt{#1}}
\providecommand{\urlprefix}{URL }
\providecommand{\doi}[1]{https://doi.org/#1}

\bibitem{anciukevicius24iclr}
Anciukevi{\v{c}}ius, T., Manhardt, F., Tombari, F., Henderson, P.: Denoising diffusion via image-based rendering. In: ICLR (2024)

\bibitem{barron2021mipnerf}
Barron, J.T., Mildenhall, B., Tancik, M., Hedman, P., Martin-Brualla, R., Srinivasan, P.P.: {Mip-NeRF}: A multiscale representation for anti-aliasing neural radiance fields. In: ICCV (2021)

\bibitem{Caron2021ICCV}
Caron, M., Touvron, H., Misra, I., J{\'{e}}gou, H., Mairal, J., Bojanowski, P., Joulin, A.: Emerging properties in self-supervised vision transformers. In: ICCV (2021)

\bibitem{charatan23pixelsplat}
Charatan, D., Li, S., Tagliasacchi, A., Sitzmann, V.: pixelsplat: 3d gaussian splats from image pairs for scalable generalizable 3d reconstruction (2024)

\bibitem{Chen2022ECCV}
Chen, A., Xu, Z., Geiger, A., Yu, J., Su, H.: {TensoRF}: Tensorial radiance fields. In: ECCV (2022)

\bibitem{Chen2023TOG}
Chen, A., Xu, Z., Wei, X., Tang, S., Su, H., Geiger, A.: Dictionary fields: Learning a neural basis decomposition. ACM Trans. Graph.  (2023)

\bibitem{Chen2023factor}
Chen, A., Xu, Z., Wei, X., Tang, S., Su, H., Geiger, A.: Factor fields: A unified framework for neural fields and beyond. arXiv.org  (2023)

\bibitem{Chen2021ICCVb}
Chen, A., Xu, Z., Zhao, F., Zhang, X., Xiang, F., Yu, J., Su, H.: {MVSNeRF}: Fast generalizable radiance field reconstruction from multi-view stereo. In: ICCV (2021)

\bibitem{chen2019point}
Chen, R., Han, S., Xu, J., Su, H.: Point-based multi-view stereo network. In: ICCV (2019)

\bibitem{chen2023matchnerf}
Chen, Y., Xu, H., Wu, Q., Zheng, C., Cham, T.J., Cai, J.: Explicit correspondence matching for generalizable neural radiance fields. arXiv.org  (2023)

\bibitem{chen2024mvsplat}
Chen, Y., Xu, H., Zheng, C., Zhuang, B., Pollefeys, M., Geiger, A., Cham, T.J., Cai, J.: Mvsplat: Efficient 3d gaussian splatting from sparse multi-view images. arXiv.org  (2024)

\bibitem{cheng2020deep}
Cheng, S., Xu, Z., Zhu, S., Li, Z., Li, L.E., Ramamoorthi, R., Su, H.: Deep stereo using adaptive thin volume representation with uncertainty awareness. In: CVPR (2020)

\bibitem{SRF}
Chibane, J., Bansal, A., Lazova, V., Pons-Moll, G.: Stereo radiance fields ({SRF}): Learning view synthesis for sparse views of novel scenes. In: CVPR (2021)

\bibitem{de1999poxels}
De~Bonet, J.S., Viola, P.: Poxels: Probabilistic voxelized volume reconstruction. In: ICCV (1999)

\bibitem{Objaverse}
Deitke, M., Schwenk, D., Salvador, J., Weihs, L., Michel, O., VanderBilt, E., Schmidt, L., Ehsani, K., Kembhavi, A., Farhadi, A.: Objaverse: {A} universe of annotated {3D} objects. In: CVPR (2023)

\bibitem{kangle2021dsnerf}
Deng, K., Liu, A., Zhu, J.Y., Ramanan, D.: Depth-supervised {NeRF}: Fewer views and faster training for free. In: CVPR (2022)

\bibitem{Transformers}
Dosovitskiy, A., Beyer, L., Kolesnikov, A., Weissenborn, D., Zhai, X., Unterthiner, T., Dehghani, M., Minderer, M., Heigold, G., Gelly, S., Uszkoreit, J., Houlsby, N.: An image is worth 16$\times$16 words: Transformers for image recognition at scale. In: ICLR (2021)

\bibitem{GSO}
Downs, L., Francis, A., Koenig, N., Kinman, B., Hickman, R., Reymann, K., McHugh, T.B., Vanhoucke, V.: Google scanned objects: {A} high-quality dataset of {3D} scanned household items. In: ICRA (2022)

\bibitem{du2023cross}
Du, Y., Smith, C., Tewari, A., Sitzmann, V.: Learning to render novel views from wide-baseline stereo pairs. In: CVPR (2023)

\bibitem{PyTorchLightning}
Falcon, W., {The PyTorch Lightning team}: {PyTorch Lightning} (2019), \url{https://github.com/Lightning-AI/lightning}

\bibitem{Keil2022CVPR}
Fridovich{-}Keil, S., Yu, A., Tancik, M., Chen, Q., Recht, B., Kanazawa, A.: Plenoxels: Radiance fields without neural networks. In: CVPR (2022)

\bibitem{furukawa2010accurate}
Furukawa, Y., Ponce, J.: Accurate, dense, and robust multiview stereopsis. PAMI  (2010)

\bibitem{Goesele2007ICCV}
Goesele, M., Snavely, N., Curless, B., Hoppe, H., Seitz, S.M.: Multi-view stereo for community photo collections. In: ICCV (2007)

\bibitem{gu2020cascade}
Gu, X., Fan, Z., Zhu, S., Dai, Z., Tan, F., Tan, P.: Cascade cost volume for high-resolution multi-view stereo and stereo matching. In: CVPR (2020)

\bibitem{esteban2004silhouette}
Hern{\'a}ndez~Esteban, C., Schmitt, F.: Silhouette and stereo fusion for {3D} object modeling. Computer Vision and Image Understanding  (2004)

\bibitem{hong2024lrm}
Hong, Y., Zhang, K., Gu, J., Bi, S., Zhou, Y., Liu, D., Liu, F., Sunkavalli, K., Bui, T., Tan, H.: {LRM}: Large reconstruction model for single image to {3D}. In: ICLR (2024)

\bibitem{HUANG2024SIGGRAPH}
Huang, B., Yu, Z., Chen, A., Geiger, A., Gao, S.: 2d gaussian splatting for geometrically accurate radiance fields. ACM SIGGRAPH  (2024)

\bibitem{im2018dpsnet}
Im, S., Jeon, H.G., Lin, S., Kweon, I.S.: {DPSNet}: End-to-end deep plane sweep stereo. In: ICLR (2019)

\bibitem{Perceiver}
Jaegle, A., Gimeno, F., Brock, A., Vinyals, O., Zisserman, A., Carreira, J.: Perceiver: General perception with iterative attention. In: Meila, M., Zhang, T. (eds.) ICML (2021)

\bibitem{GeoNeRF}
Johari, M.M., Lepoittevin, Y., Fleuret, F.: {GeoNeRF}: Generalizing {NeRF} with geometry priors. In: CVPR (2022)

\bibitem{Karras2019CVPR}
Karras, T., Laine, S., Aila, T.: A style-based generator architecture for generative adversarial networks. In: CVPR (2019)

\bibitem{kerbl3Dgaussians}
Kerbl, B., Kopanas, G., Leimk{\"u}hler, T., Drettakis, G.: {3D Gaussian} splatting for real-time radiance field rendering. ACM Trans. on Graphics  (2023)

\bibitem{kolmogorov2002multi}
Kolmogorov, V., Zabih, R.: Multi-camera scene reconstruction via graph cuts. In: ECCV (2002)

\bibitem{kulhanek2022viewformer}
Kulh{\'a}nek, J., Derner, E., Sattler, T., Babu{\v{s}}ka, R.: {ViewFormer}: {NeRF}-free neural rendering from few images using transformers. In: ECCV (2022)

\bibitem{kutulakos2000theory}
Kutulakos, K.N., Seitz, S.M.: A theory of shape by space carving. International journal of computer vision  (2000)

\bibitem{li2024instantd}
Li, J., Tan, H., Zhang, K., Xu, Z., Luan, F., Xu, Y., Hong, Y., Sunkavalli, K., Shakhnarovich, G., Bi, S.: {Instant3D}: Fast text-to-{3D} with sparse-view generation and large reconstruction model. In: ICLR (2024)

\bibitem{ENeRF}
Lin, H., Peng, S., Xu, Z., Yan, Y., Shuai, Q., Bao, H., Zhou, X.: Efficient neural radiance fields for interactive free-viewpoint video. In: SIGGRAPH Asia (2022)

\bibitem{liu2023zero1to3}
Liu, R., Wu, R., Hoorick, B.V., Tokmakov, P., Zakharov, S., Vondrick, C.: Zero-1-to-3: Zero-shot one image to {3D} object. In: ICCV (2023)

\bibitem{long2023wonder3d}
Long, X., Guo, Y.C., Lin, C., Liu, Y., Dou, Z., Liu, L., Ma, Y., Zhang, S.H., Habermann, M., Theobalt, C., et~al.: {Wonder3D}: Single image to {3D} using cross-domain diffusion. arXiv.org  (2023)

\bibitem{Loshchilov2019ICLR}
Loshchilov, I., Hutter, F.: Decoupled weight decay regularization. In: ICLR (2019)

\bibitem{luo2019p}
Luo, K., Guan, T., Ju, L., Huang, H., Luo, Y.: {P-MVSNet}: Learning patch-wise matching confidence aggregation for multi-view stereo. In: ICCV (2019)

\bibitem{Mildenhall2020ECCV}
Mildenhall, B., Srinivasan, P.P., Tancik, M., Barron, J.T., Ramamoorthi, R., Ng, R.: {NeRF}: Representing scenes as neural radiance fields for view synthesis. In: ECCV (2020)

\bibitem{Miyato2024GTA}
Miyato, T., Jaeger, B., Welling, M., Geiger, A.: {GTA}: A geometry-aware attention mechanism for multi-view transformers. In: ICLR (2024)

\bibitem{Mueller2022TOG}
M\"uller, T., Evans, A., Schied, C., Keller, A.: Instant neural graphics primitives with a multiresolution hash encoding. ACM Trans. on Graphics  (2022)

\bibitem{Nguyen-Phuoc2019ICCV}
Nguyen{-}Phuoc, T., Li, C., Theis, L., Richardt, C., Yang, Y.: {HoloGAN}: Unsupervised learning of {3D} representations from natural images. In: ICCV (2019)

\bibitem{Niemeyer2022CVPR}
Niemeyer, M., Barron, J., Mildenhall, B., Sajjadi, M.S.M., Geiger, A., Radwan, N.: {RegNeRF}: Regularizing neural radiance fields for view synthesis from sparse inputs. In: CVPR (2022)

\bibitem{Niemeyer2020CVPR}
Niemeyer, M., Mescheder, L., Oechsle, M., Geiger, A.: Differentiable volumetric rendering: Learning implicit {3D} representations without {3D} supervision. In: CVPR (2020)

\bibitem{DiT}
Peebles, W., Xie, S.: Scalable diffusion models with transformers. In: ICCV (2023)

\bibitem{qiu2024richdreamer}
Qiu, L., Chen, G., Gu, X., Zuo, Q., Xu, M., Wu, Y., Yuan, W., Dong, Z., Bo, L., Han, X.: Richdreamer: A generalizable normal-depth diffusion model for detail richness in text-to-3d. In: CVPR (2024)

\bibitem{reizenstein21co3d}
Reizenstein, J., Shapovalov, R., Henzler, P., Sbordone, L., Labatut, P., Novotny, D.: Common objects in {3D}: Large-scale learning and evaluation of real-life {3D} category reconstruction. In: ICCV (2021)

\bibitem{schonberger2016pixelwise}
Sch{\"o}nberger, J.L., Zheng, E., Frahm, J.M., Pollefeys, M.: Pixelwise view selection for unstructured multi-view stereo. In: ECCV (2016)

\bibitem{Schoenberger2016CVPR}
Schönberger, J.L., Frahm, J.M.: Structure-from-motion revisited. In: CVPR (2016)

\bibitem{seitz2006comparison}
Seitz, S.M., Curless, B., Diebel, J., Scharstein, D., Szeliski, R.: A comparison and evaluation of multi-view stereo reconstruction algorithms. In: CVPR (2006)

\bibitem{shi2023zerorf}
Shi, R., Wei, X., Wang, C., Su, H.: {ZeroRF}: Fast sparse view 360° reconstruction with zero pretraining (2023), arXiv:2312.09249

\bibitem{shi2023MVDream}
Shi, Y., Wang, P., Ye, J., Mai, L., Li, K., Yang, X.: {MVDream}: Multi-view diffusion for {3D} generation. In: ICLR (2024)

\bibitem{NOAH2006TOG}
Snavely, N., Seitz, S.M., Szeliski, R.: Photo tourism: exploring photo collections in {3D}. ACM Trans. on Graphics  (2006)

\bibitem{somraj2022decompnet}
Somraj, N., Karanayil, A., Soundararajan, R.: {SimpleNeRF}: Regularizing sparse input neural radiance fields with simpler solutions. In: SIGGRAPH Asia (2023)

\bibitem{SunSC22}
Sun, C., Sun, M., Chen, H.: Direct voxel grid optimization: Super-fast convergence for radiance fields reconstruction. In: CVPR (2022)

\bibitem{szymanowicz2024splatter_image}
Szymanowicz, S., Rupprecht, C., Vedaldi, A.: Splatter image: Ultra-fast single-view 3d reconstruction. CVPR  (2024)

\bibitem{tang2024lgm}
Tang, J., Chen, Z., Chen, X., Wang, T., Zeng, G., Liu, Z.: {LGM}: Large multi-view {Gaussian} model for high-resolution {3D} content creation. arXiv.org  (2024)

\bibitem{sparf2023}
Truong, P., Rakotosaona, M.J., Manhardt, F., Tombari, F.: {SPARF}: Neural radiance fields from sparse and noisy poses. In: CVPR (2023)

\bibitem{Vaswani2017NIPS}
Vaswani, A., Shazeer, N., Parmar, N., Uszkoreit, J., Jones, L., Gomez, A.N., Kaiser, L., Polosukhin, I.: Attention is all you need. In: NIPS. pp. 5998--6008 (2017)

\bibitem{venkat2023geometry}
Venkat, N., Agarwal, M., Singh, M., Tulsiani, S.: Geometry-biased transformers for novel view synthesis. arXiv.org  (2023)

\bibitem{SparseNeRF}
Wang, G., Chen, Z., Loy, C.C., Liu, Z.: {SparseNeRF}: Distilling depth ranking for few-shot novel view synthesis. In: ICCV (2023)

\bibitem{Wang2021NEURIPSa}
Wang, P., Liu, L., Liu, Y., Theobalt, C., Komura, T., Wang, W.: {NeuS}: Learning neural implicit surfaces by volume rendering for multi-view reconstruction. In: NeurIPS (2021)

\bibitem{wang2024pflrm}
Wang, P., Tan, H., Bi, S., Xu, Y., Luan, F., Sunkavalli, K., Wang, W., Xu, Z., Zhang, K.: {PF}-{LRM}: Pose-free large reconstruction model for joint pose and shape prediction. In: ICLR (2024)

\bibitem{wang2021ibrnet}
Wang, Q., Wang, Z., Genova, K., Srinivasan, P., Zhou, H., Barron, J.T., Martin-Brualla, R., Snavely, N., Funkhouser, T.: {IBRNet}: Learning multi-view image-based rendering. In: CVPR (2021)

\bibitem{dust3r2023}
Wang, S., Leroy, V., Cabon, Y., Chidlovskii, B., Jerome, R.: Dust3r: Geometric 3d vision made easy. CVPR  (2024)

\bibitem{wu2023reconfusion}
Wu, R., Mildenhall, B., Henzler, P., Park, K., Gao, R., Watson, D., Srinivasan, P.P., Verbin, D., Barron, J.T., Poole, B., Holynski, A.: {ReconFusion}: {3D} reconstruction with diffusion priors (2023), arXiv:2312.02981

\bibitem{gobjaverse}
Xu, C., Dong, Y., Zuo, Q., Zhang, J., Ye, X., Geng, W., Zhang, Y., Gu, X., Qiu, L., Zhao, Z., Qing, R., Jiayi, J., Dong, Z., Bo, L.: G-buffer {Objaverse}: High-quality rendering dataset of {Objaverse}, \url{https://aigc3d.github.io/gobjaverse/}

\bibitem{xu2023murf}
Xu, H., Chen, A., Chen, Y., Sakaridis, C., Zhang, Y., Pollefeys, M., Geiger, A., Yu, F.: {MuRF}: Multi-baseline radiance fields. In: CVPR (2024)

\bibitem{xu2024dmvd}
Xu, Y., Tan, H., Luan, F., Bi, S., Wang, P., Li, J., Shi, Z., Sunkavalli, K., Wetzstein, G., Xu, Z., Zhang, K.: {DMV3D}: Denoising multi-view diffusion using {3D} large reconstruction model. In: ICLR (2024)

\bibitem{yao2018mvsnet}
Yao, Y., Luo, Z., Li, S., Fang, T., Quan, L.: {MVSNet}: Depth inference for unstructured multi-view stereo. In: ECCV (2018)

\bibitem{yao2019recurrent}
Yao, Y., Luo, Z., Li, S., Shen, T., Fang, T., Quan, L.: Recurrent {MVSNet} for high-resolution multi-view stereo depth inference. In: CVPR (2019)

\bibitem{Yariv2020NIPS}
Yariv, L., Kasten, Y., Moran, D., Galun, M., Atzmon, M., Ronen, B., Lipman, Y.: Multiview neural surface reconstruction by disentangling geometry and appearance. In: NIPS (2020)

\bibitem{xu2024grm}
Yinghao, X., Zifan, S., Wang, Y., Hansheng, C., Ceyuan, Y., Sida, P., Yujun, S., Gordon, W.: Grm: Large gaussian reconstruction model for efficient 3d reconstruction and generation (2024)

\bibitem{Yu2021CVPR}
Yu, A., Ye, V., Tancik, M., Kanazawa, A.: {pixelNeRF}: Neural radiance fields from one or few images. In: CVPR (2021)

\bibitem{Yu2023MipSplatting}
Yu, Z., Chen, A., Huang, B., Sattler, T., Geiger, A.: Mip-splatting: Alias-free 3d gaussian splatting. CVPR  (2024)

\end{thebibliography}

\appendix

\section{Limitations and Discussions} 
Our \emph{LaRa} demonstrates a remarkable efficiency feed-forward model that achieved high-fidelity all-around novel-view synthesis and surface reconstruction from sparse large-baseline images.
However, our approach struggles to recover high-frequency geometry and texture details, mainly due to the low volume resolution.
Enhancing our approach with techniques such as gradient checkpointing or mixed-precision training can potentially increase training batch size as well as volume resolution.
We have also noticed that our method can yield inconsistent rendering results when the geometry is incorrectly estimated or when reconstructing multi-view inconsistent inputs, as demonstrated in the comparison video. This occurs because our method utilizes second-order Spherical Harmonic appearance modeling. Although such modeling can capture view-dependent effects, it also introduces a stronger ambiguity between geometry and appearance. We believe that incorporating our method with a physically-based rendering process can potentially address this issue.
In addition, our work assumes posed inputs, but estimating precise camera poses for sparse views is a challenge in practice.
Incorporating a pose estimation module \cite{wang2024pflrm} into the feed-forward setting is an orthogonal direction to our work.

\section{Acknowledgements}
We thank Bozidar Antic for pointing out a bug that resulted in an improvement of about 1dB. 
Special thanks to BinBin Huang and Zehao Yu for their helpful discussion and suggestions.
We would like to thank Bi Sai, Jiahao Li, Zexiang Xu for providing us with the testing examples of Instant3D, and Jiaxiang Tang for helping us to construct a comparison with LGM. This project was supported by the ERC Starting Grant LEGO-3D (850533) and the DFG EXC number 2064/1 - project number 390727645.

\section{More Visual Results}
\begin{figure}[t]
    \includegraphics[width=\linewidth]{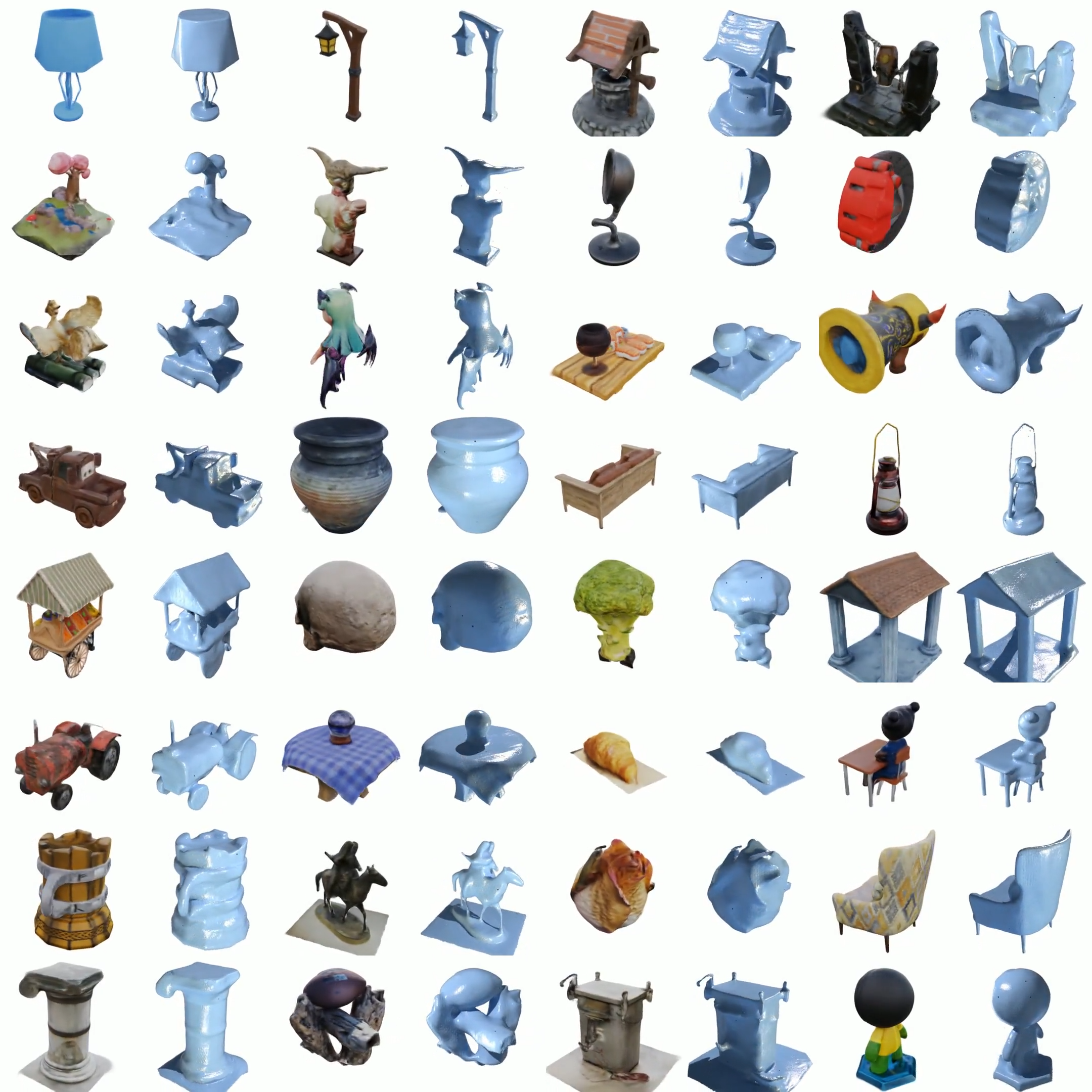}
    \caption{Reconstruction results on Gobjaverse testing set.}
    \label{fig:gobjaverse}
\end{figure}

\begin{figure}[t]
    \includegraphics[width=\linewidth]{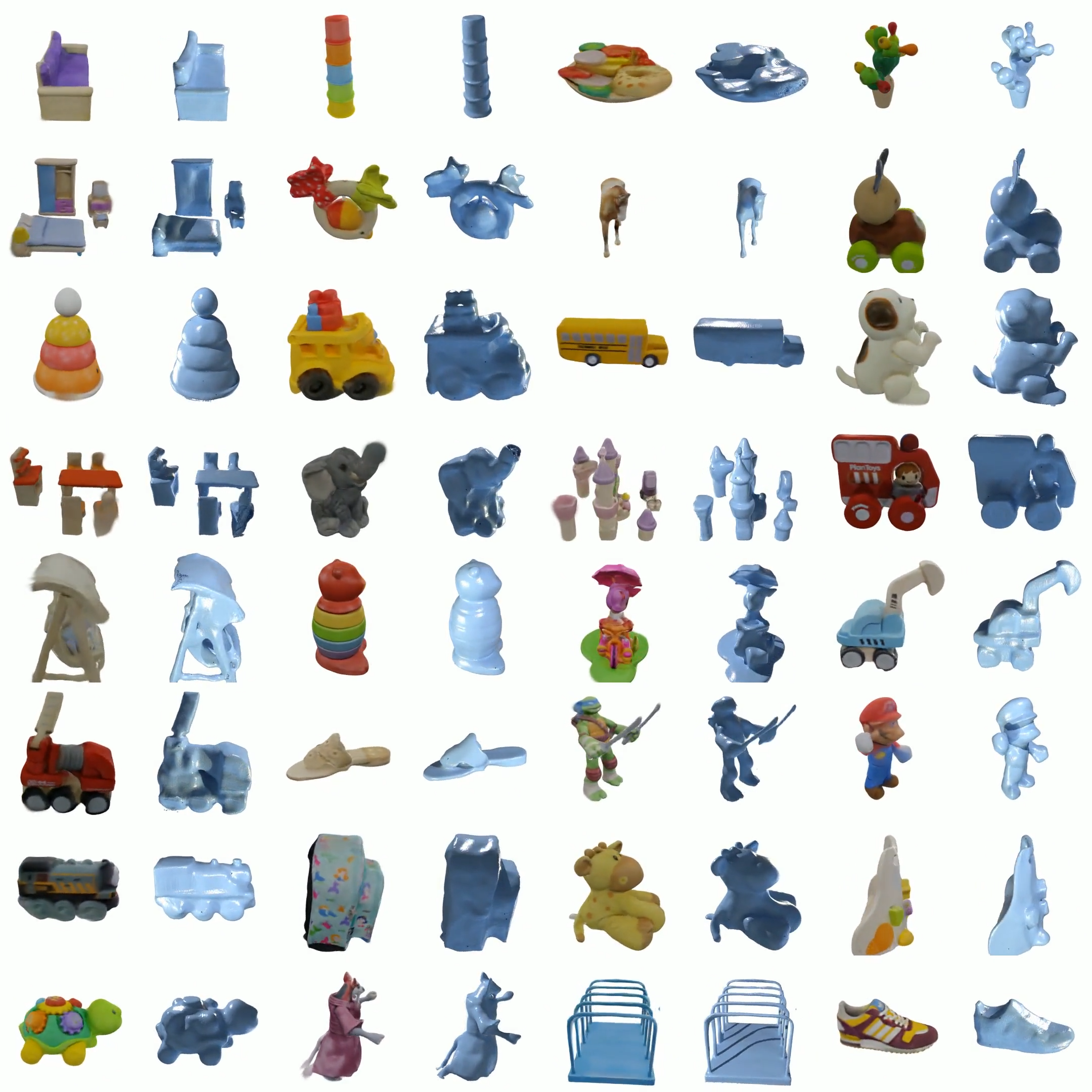}
    \caption{Reconstruction results on Google Scanned Object dataset.}
    \label{fig:gso}
\end{figure}

\begin{figure}[t]
    \includegraphics[width=\linewidth]{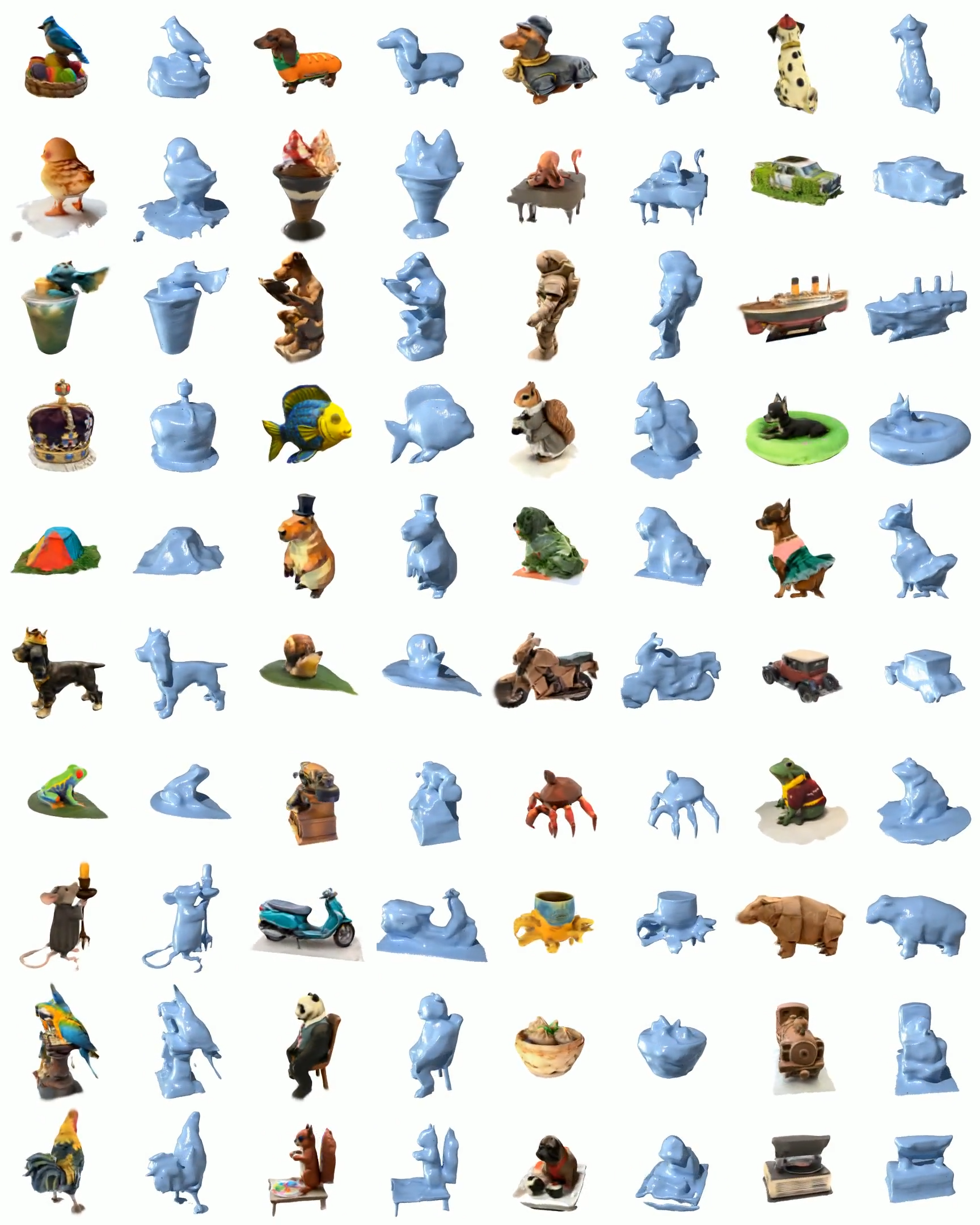}
    \caption{Reconstruction results on Instant3D scenes.}
    \label{fig:instant3d}
\end{figure}

Finally, we report more qualitative results on the testing sets, as shown in \figref{fig:gobjaverse},\figref{fig:gso}, and \figref{fig:instant3d}.

\end{document}